\documentclass[conference]{IEEEtran}
\IEEEoverridecommandlockouts
\usepackage{cite}
\usepackage{amsmath,amssymb,amsfonts}
\usepackage{algorithmic}
\usepackage{graphicx}
\usepackage{textcomp}
\usepackage{xcolor}
\usepackage{listings}
\usepackage{float}
\usepackage{mathtools, cuted}
\usepackage{lipsum}
\usepackage{tikz}

\begin{document}

\IEEEoverridecommandlockouts
\IEEEpubid{\makebox[\columnwidth]{978-1-5386-5541-2/18/\$31.00~\copyright2018 IEEE \hfill} \hspace{\columnsep}\makebox[\columnwidth]{ }}



\title{Automatic difficulty management and testing in games using a framework based on behavior trees and genetic algorithms\\
}

\author{\IEEEauthorblockN{Ciprian Paduraru}
\IEEEauthorblockA{\textit{The Research Institute of the }\\
\textit{University of Bucharest (ICUB), Romania}\\ 
\textit{Department of Computer Science,}\\
\textit{University of Bucharest, Romania} \\
ciprian.paduraru@fmi.unibuc.ro}
\and
\IEEEauthorblockN{Miruna Paduraru}
\IEEEauthorblockA{\textit{Ubisoft Entertainment and } \\
\textit{Department of Computer Science,}\\
\textit{University of Bucharest, Romania} \\
miruna-gabriela.paduraru@ubisoft.com}
}

\maketitle

\IEEEpubidadjcol

\begin{abstract}
The diversity of agent behaviors is an important topic for the quality of video games and virtual environments in general. Offering the most compelling experience for users with different skills is a difficult task, and usually needs important manual human effort for tuning existing code. This can get even harder when dealing with adaptive difficulty systems. Our paper’s main purpose is to create a framework that can automatically create behaviors for game agents of different difficulty classes and enough diversity. In parallel with this, a second purpose is to create more automated tests for showing defects in the source code or possible logic exploits with less human effort.
\end{abstract}

\IEEEpubidadjcol


\IEEEpubidadjcol

\section{Introduction} \label{sec:introduction}
Behavior trees (BT) are one of the core tools used today in the decision-making layer of simulation softwares (e.g. video games, crowd simulations, etc), and recently attracted the robotics field. Agents can leverage on behavior trees to take decisions considering both distributed and centralized architectures.  
As the related work section describes from state-of-the-art, the main advantages of behavior trees over other decision making methods are: 

\begin{itemize} 
\item Increased Modularity - a key property to enable reusability of code, incremental design of functionality and efficient testing of functionality. Behavior trees provide more flexibility, easier to understand, prototype and maintain when implementing complex behaviors.
\item BTs have more powerful semantics than other systems. A behavior tree can actually be seen as a programming language with function calls, tests or loops.
\end{itemize}

From our practical experience in video games, difficulty tuning is a very challenging task. The usual workflow is to create first a good AI which poses challenges for most of the users. Then, developers have to think about how to change the source code and behavior parameters of the AI system to produce the right level of challenge depending on the user's skills - which is usually specified in a game by selecting a difficulty setting such as \textit{Easy, Medium, or Hard}. The question is, which behaviors should be available per each class of difficulty? Or, which parameters should be used for each difficulty class and behavior? In general, this ends up with a lot of manual work that makes the source code implementation of AI systems full of magic constant values and branch conditions depending on difficulty settings. This manual work is time consuming and error prone, since the feedback in general is collected from humans testing the game. Also, it would be ideal to offer the right level of challenge for each user. What if users would learn the game very quickly and the pre-defined \textit{Hard} difficulty class would not be enough for them anymore? This basically requires adaptive difficulty and it is impossible to offer it with manual tuning the code since developers can't release a different game product for each client.

Another observation is that sometimes users can find a way to obtain important rewards by observing some shortcuts in the logic of a game (also known as \textit{exploits}). Finding such vulnerabilities for the game's ecosystem is again time consuming and needs human observation. 
Testing the source code against defects is also time consuming in general, since in gaming, it is done by humans. While it is true that most of the game components such as rendering or animation systems need some level of human observation, we think that the implementations of AI systems inside a game can be automated more to reduce human effort and costs.

Finally, we also observed that with scripted behaviors, even when changing and randomizing parameters around, the users can discover the patterns behind a game's AI quickly. One way to fix this would be to create more diversity in terms of behaviors too, not only behaviors' parameters.

Our work concentrates on solving the issues described above. At the moment of writing this paper, this is the first study that addresses the problem of using behavior trees for automatic testing and difficulty tuning in games. The main contribution of the paper is to report and sketch a framework implementation that uses behavior trees and genetic algorithms at its core, which could provide the following features to the game development industry:

\begin{itemize}
\item Automatic difficulty management in games. Difficulty can also be adapted per user to offer the right level of challenge for each level.
\item Create more diversity by generating new behaviors with less human effort.
\item Automatically identify possible exploits that users can use in their advantage.
\item A way to automatize more the functional testing of the gameplay components.
\end{itemize}

Also, compared to the current state of the art, the paper provides an extended analysis of the mutation options available for using genetic algorithms on behavior trees, and a novel method for similar behaviors elimination. The mentioned features were evaluated using a 3D game environment used for prototyping and quick results evaluation created by us and made open source for the community.

The rest of the paper is organized as follows. Section \ref{sec:relatedWork} presents the state of the art of using Behavior Trees in different areas, with an emphasis over its applications inside game development and robotics. Section \ref{sec:btbasics} is an introduction for readers over basic concepts related to Behavior Trees. The open-source framework that we've built on top of this concept for providing automatic difficulty tuning and testing is described in Section \ref{sec:optimizations}. Finally, evaluation results are shown in Section \ref{sec:evaluation}, while the last section draws the conclusion and presents our future work plans.

\section{Related work} \label{sec:relatedWork}
BTs have been extensively used in robotic manipulation in \cite{DBLP:conf/iros/BagnellCCGHKKLLPPVZ12}, \cite{15}. One afirmation that sustains the modularity advantage of BTs can be understood from the following quote ``The main advantage is that individual behaviors can easily be reused in the context of another higher level behavior, without needing to specify how they relate to subsequent behaviors'' \cite{DBLP:conf/iros/BagnellCCGHKKLLPPVZ12}.

In \cite{DBLP:journals/trob/ColledanchiseO17} authors are using behavior trees to increase the modularity of the Hybrid Control Systems and prove that they are the generalization of three classical concepts from the robot control literature: Subsumption architecture \cite{1087032}, Sequential behavior compositions \cite{20}, and Decision trees \cite{DBLP:journals/trob/ColledanchiseO17}.
BTs have been even suggested as a key component in brain surgery robotics in \cite{7139738} due to their modularity and simple syntax. Also, \cite{7140065} notes that they can be used by non-experts to pick-up implemented nodes and tasks and just create a behavior. This is also sustained by the video game industry which uses non-technical people to create behaviors of agents by using a library of nodes. A unified behavior tree framework for robot control that is applied on a NAO robot is described in \cite{6907656}.

Some machine learning applications oriented to video games development, such as \cite{DBLP:conf/evoW/Colton12} and \cite{7435292}, are relying on BTs. Eficient parametrization of BTs was investigated in \cite{Shoulson:2011:PBT:2177817.2177835}. Formal verification in navigation mission plans is studied in \cite{13}.

Behavior Trees have been also successfully applied in evolving agents’ behaviors in games or applications oriented for visual simulations. In \cite{article1}, the authors proposed a model-free framework as against model-based framework; they started from an initial state with a BT that consists of only one node, which is an action node chosen using a greedy search process. In the case that such an action is not found, their Genetic Programming process will be initiated with a population of binary trees with random node assignments consisting of a combination of conditions and action nodes. They used crossover, mutation and selection as genetic operators. To identify the redundant or unnecessary sub-trees they used anti-bloat control technique. In another interesting example in \cite{article2}, the authors applied Grammatical Evolution to generate different levels of environment for Mario game. A BT’s root node is a selector with a variable number of Behavior Block (BB) subtrees, encoding sub-behaviors. The four BB of the best BT generated was chosen and an agent using it was sent to the Mario AI game simulation. In \cite{article4}, the author presents bots for Zero-K game – a real time strategy game. The objective of the thesis was to write different types of BTs with different characteristics, to make the game more varied in terms of AI behaviors. In \cite{article5}, authors used BTs to design and develop a competitive player which was able to outperform the DEFCON game’s original AI-bot more than $50\%$ of the time. An important take away from this paper is that authors are using clusters of BTs for different aspects of their games (e.g. economy, defense, etc) since it was simpler to obtain good behaviors this way. Our current work reuses the genetic operations and previous ideas, extending the state-of-the-art with the objectives mentioned in Section 1.

\section{Behavior trees concept} \label{sec:btbasics}
This section gives a practically oriented description of behavior trees. A more formal specification can be found in \cite{6907656} and \cite{DBLP:journals/trob/ColledanchiseO17}. Also, a quick comparison with machine learning techniques is presented at the end of this section.

\subsection{Nodes semantics and examples}

A BT is a directed tree with the usual definition of nodes, edges, children, parents, and leaves. The internal nodes are used to control the logical flow, while the leaves execute actions or test different conditions (e.g. sensing operations or movement). The evaluation starts from the root node which is \textit{ticked} with a given frequency ($Delta_t$) and then propagates following the rules defined in the internal nodes until it reaches a leaf node. A node can return either \textit{Success}, \textit{Failure}, or \textit{Running}. For instance, a node performing pathfinding would return \textit{Success} when the agent arrived at the target, \textit{Failure} when blocked and no path can be computed, and \textit{Running} when the agent is still performing movements to find the target without being blocked. The return value from a leaf is then propagated back to the root of the tree with different rules depending on each category of internal node. These rules are described in the text below. 

\begin{figure}[t]
\begin{center}
\includegraphics[width=9cm]{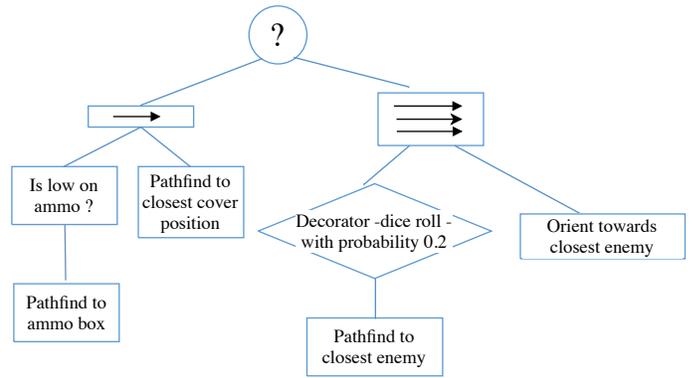}
\end{center}
\caption{A behavior tree example for an agent representing a tank.}\label{fig:FullTree}
\end{figure}

\textit{Selectors}. A selector node (question mark in Figure \ref{fig:FullTree}) is used to represent alternatives of actions. Each child represents one of these alternatives and in order to succeed, only one of them must succeed. The children are evaluated from left to right in deterministic Selectors, but can also be evaluated in non-deterministic order for a more variety of actions (e.g. have agents that behave differently between ticks). If one of the children returns \textit{Running}, the tick of this node stops and the return value is sent to its parent. 

\textit{Sequence}. A sequence node (simple arrow mark in Figure \ref{fig:FullTree}) is used to represent a sequence of actions that an agent must do. The difference between Selector and Sequence is that the latter, needs to have all children evaluated as Success in order to return Success further. It is also possible to make this node non-deterministic and execute the sequence of actions in a different way each time, when the order of actions does not matter. 

\textit{Parallel}. A parallel node (the three arrows mark in Figure \ref{fig:FullTree}) ticks all its children at the same time. Considering that the number of nodes which returns success is represented by $C$, $N$ total children, and a parameter $K$ set by user, then the parallel node returns success if $C \geq K$, failure if $C < N-K$, and running otherwise. Altough its name could mean parallelization, it is not necessary to execute the children in parallel in the backend. The role of this type of node is to allow different children run in the same tick (e.g. performing at the same time: movement, speak and fire).

\textit{Action} and {Condition}. These are leaves node performing actions such as movements or checking environment. Can be used to mutate the state of the agent running it or send messages to other agents.

\textit{Decorators}. A decorator has a single child and modifies its behavior in some way. A simple case is inverting the result of the child (e.g. switch Failure with Success). Other cases can include running a node for a limited wall-time or a number of iterations (set by user) until it returns Success.

\textit{ProxyNodes}. A proxy node is used to make a reference to another behavior tree. This is used to reuse existing functionalities.

An example of a tree with the nodes explained above can be found in Figure \ref{fig:FullTree}. The root of the BT is a selector node (question mark symbol) with two children. The first child will make the agent find an ammo box when the ammo is considered low, followed by looking for a cover position by using a sequence node (the arrow symbol). The right child is able to perform two actions in the same tick by using a parallel node (the multiple arrow symbol). The decorator, in this case, is used to go and search for an enemy with probability $0.2$ on each tick. At the same time, the tank is oriented towards the closest enemy, such that the agent is prepared to fire when an enemy is in sight, even when pathfinding is used. The leaves of the tree are \textit{Action} nodes in this case, but often to simplify visualization and promote reusability, leaves are \textit{ProxyNodes} defining complex actions. This way, modular hierarchical behaviors can be made using behavior trees' semantics.

\subsection{High level usage of behavior trees and communication} 

Each game agent (which in our simulation environment is a tank game object) has attached a behavior tree controller. The concept of \textit{blackboards} is widely used to communicate between AI systems \cite{DBLP:journals/jirs/ShinKK18}. In our implementation of behavior trees, there are two kinds of blackboards:
\begin{itemize}
\item \textit{Global blackboard}: Contains information exposed by the game and being shared (public state) by all agents in the simulation. E.g. the location of agents, the location of upgrade boxes, the health status of each agent, etc.
\item \textit{Local blackboard}: It is local per agent and contains the state of its own AI system (private state). For instance, if a pathfinding node is in progress, it retains that state inside its local blackboard. Another example are decorator nodes, which sometimes need persistent state between successive executions.
\end{itemize}

\subsection{Quick comparison between classic decision-making techniques and machine learning}
Due to space limits, only some brief ideas are sketched in this subsection, without any formalism or demonstrated evaluation. From our practical experience in applying machine learning models for game development, we concluded that they have the following two advantages in comparison with having human writing scripted behaviors (including behavior trees):
\begin{enumerate}
\item{In general, by writing scripted behaviors, it is difficult and time-consuming to match the same level of experience offered by using instead machine learning techniques.}
\item{Many times we found that scripted behaviors are taking more execution time than inferencing a machine learning model which solves the same problem.} 
\end{enumerate}

However, the main disadvantage of machine learning models for game development is the lack of control when applying constraints in dynamic environments. As a concrete example, think of a game where the initial design changes and the previously learned model should now be used only under a certain set of conditions (possibly stochastics). 

We are not advocating for one or another, but we think that in game development, blending between classical decision making methods and machine learning ones is one of the most efficient ways to get optimal results at the moment. For behavior trees, this blend means having the action nodes perform inference on machine learning models, while the needed input and output model data reside in the blackboards.

\section{Framework implementation and usage} \label{sec:optimizations}
This section presents the architecture and the technical aspects behind our framework implementation. Our target was to make it independent of the application type (either a game or a virtual environment simulation application), so it is expected for readers to reuse our framework for many kinds of applications they need with small efforts. The source code and our test environment is open-source and available here $https://github.com/AGAPIA/BTreeGeneticFramework$. 

The test environment is based on assets imported from Unity store \cite{UnityT}. It is a 3D (isometric view) tanks game where N agents (each one representing a tank instance) compete against each other in consecutive rounds, with the purpose of defeating all others. Agents have a limited number of lives, health, and ammo. They can fire projectiles which can damage others and subtract their health value. Random boxes appear from time to time on the ground level (uniformly distributed on the available walkable space) and agents can get them to receive: \textit{shield} (not being affected by projectiles for a limited time), \textit{weapon upgrade} (increased damage is taken by the weapon), \textit{ammo} and \textit{health refill}. An image with the simulated environment is presented in Figure \ref{fig:capture}. Each agent receives a different unique behavior tree instance, whereby instance we mean both structure and parameters.

\subsection{Interaction with the game application}

\begin{figure*}[t]
\begin{center}
\includegraphics[width=14cm]{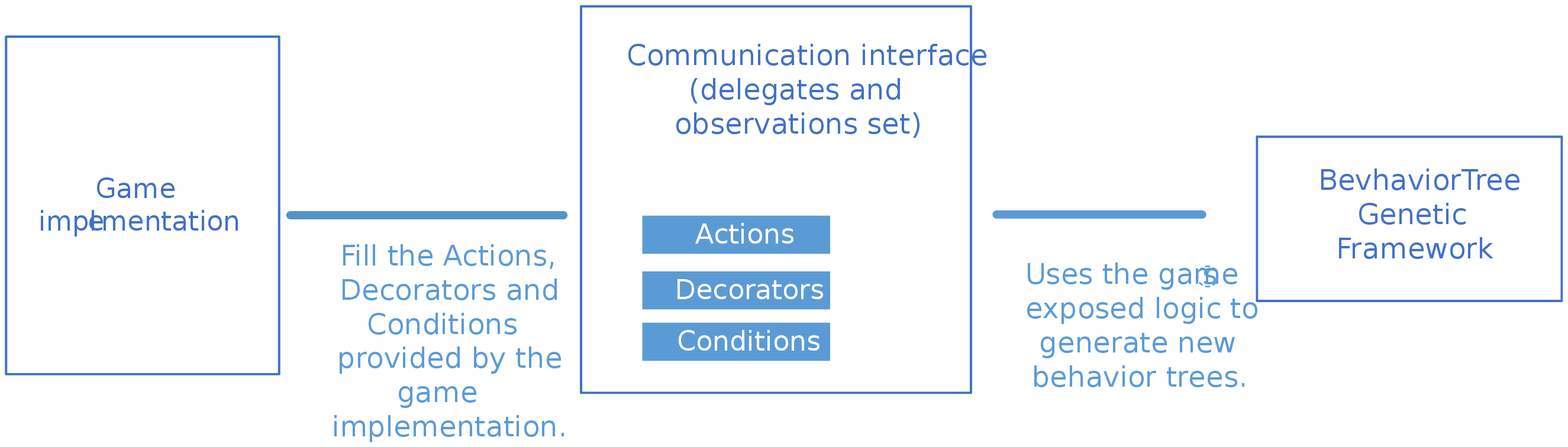}
\end{center}
\caption{Connection architecture between a game application and our framework. The purpose is to decouple the two applications such that the only dependencies are based on abstract interfaces.}\label{fig:communication}
\end{figure*}

The game using our framework must expose a certain interface in order to create behavior trees at runtime. An object named \textit{Communication interface} (Figure \ref{fig:communication}) with the following delegates must be provided by the game to the framework:

\begin{itemize}
\item \textbf{Actions} Set: a collection of actions available and implemented in the game application. Example from our environment: go to a certain position on the map, fire forward, fire ahead in enemy's movement direction, pathfind to the closest enemy, move with back, pathfind to the closest upgrade box, etc.
\item \textbf{Conditions}: a collection of things that can be tested from the environment and added to the tree to execute or not certain branches. Examples in our environment: is there any enemy in view? Am I low on ammo or health? What is the health or ammo of an enemy? Does enemy possess shield or weapon upgrade? etc.
\item \textbf{Decorators}: these look like conditions but are not related to the game environment (as the above section is), instead they are more using statistics about an agent's behavior tree execution state. These can be added especially for creating diversity. In our environment, we used decorators such as: When was this tree branch executed last time? chance for executing a branch (i.e. dice roll if a certain branch should be executed or not), Should I randomize the order of nodes below (to create diversity for selector nodes and get each time a different order of events.)? 
\end{itemize}

All of the above can become nodes in the behavior trees generated by the framework. Without going too deep in implementation details, we mention that conditions are implemented like simple expression trees with variables that can be found either in the global or local agent's blackboard. At runtime, the tree is evaluated, global or local blackboards are accessed, and the root returns the result of the condition. The decorator nodes are reading / writing data to the local blackboard of the agent. For instance, if a decorator node conditioning the time rate at which its children evaluation goes through is used, then, at each new branch evaluation the agent's local blackboard is read to test the last execution time, and at each new execution the local blackboard is written with the current tick. Writing to local blackboards can also happen inside actions evaluation. Actions must be implemented in the game application by reading data exposed only to either global or local blackboard. Each of these readings and writings are binded at initialization time, so each operation is done with $O(1)$ time complexity.

The \textit{Game simulation handler} is a component that allows the framework to test the behavior trees performance by interacting with a game level and collecting back statistics about agents. \textit{Difficulty settings specifications} is a descriptive set of metrics that must be accomplished on average by an agent to be considered in certain classes of difficulties. A detailed description of this component is presented in \ref{sec:difficulty}.

\subsection{Using parameters to create more diversity}

The set of conditions and decorators above can be parametrized to create diversity. They are actually used as templates by the framework generating new behaviors. For example, the condition node ``Am I low on ammo or health'' could have different thresholds between behavior trees using the same template nodes. For instance, one agent could consider having a low health state if the percent is below $15$, while another below $25$. Same is valid for decorators. As an example, consider having a dice rolling decorator node on a branch performing pathfinding to an enemy when there is nothing in agent's sight. Some agents could be defensive and thus, set a low probability there, while others can become more aggressive by setting a higher probability value.

\subsection{Genetic algorithm operations} \label{sec:rules}

The internal genetic algorithm implementation for generating new behavior trees is using common operations presented in the state-of-the-art work such as mutation, crossover and selection. 

The initialization module generates a configurable number $N$ of initial individuals in a genetic population $G$, where an individual is a behavior tree.
The framework can create a population of behavior trees from scratch without human effort (using randomizations with the same rules as presented below in Mutation operator). However, for faster converge as the evaluation section shows, it is still recommended to start with a small library of examples created by hand.

\begin{figure*}[t]
\begin{center}
\includegraphics[width=18cm]{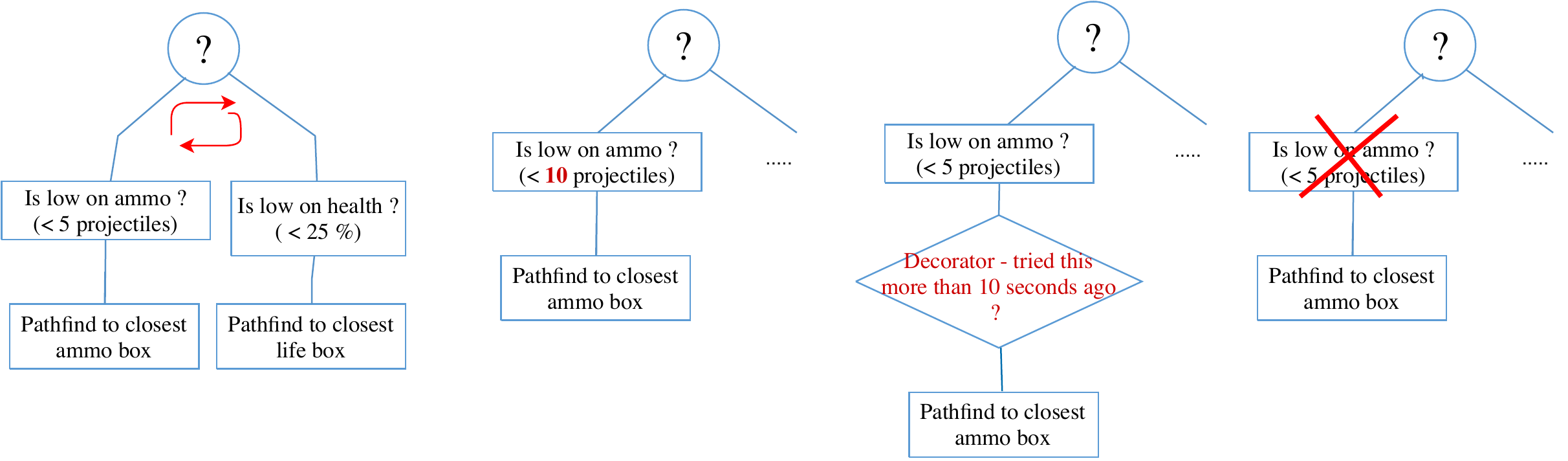}
\end{center}
\caption{Examples of how a tree structure or parameters can be mutated. In the left side, there is a shuffling operation done over a selector node, to change the priorities of the children nodes. Following, the figure shows operations for mutating a parameter, an existing tree edge by adding a decorator node or eliminating a node and its entire subtree.}\label{fig:mutation}
\end{figure*}

\textbf{Mutation}
Mutation can be done at multiple levels inside our framework (Figure \ref{fig:mutation}). A single operation from the following list will be used to mutate.

\begin{enumerate}

\item{\textit{Parameters level:}} For each type of parameter specified in the communication interface (Figure \ref{fig:communication}), the user can select a range of values (min-max or discrete) that are considered correct. The mutation will be done with operations such as bit/byte flipping, but still satisfying the user's constraints.

\item{\textit{New nodes added:}} 
At each mutation, a single node can be added to the tree. Since leaves are usually actions in a behavior tree, the new nodes must be appended to internal nodes. The rules we used to append new nodes are the following:
\begin{itemize}
\item Below an existing sequence or selector a new action can be added anywhere in the children list.
\item On a tree edge connecting a sequence or selector node and any other node, a new decorator or condition node can be added between parent and child.
\end{itemize}

\item{\textit{Changing semantics of existing nodes:}} More specifically, a selector node can be changed to a sequence and inverse.

\item{\textit{Deleting existing nodes:}} 
Every node excepting the root can be deleted. However, when a node is deleted, a check must be done to test if the deleted node is the only child in its parent, and recursively delete the parent if it is the case. Note that the entire subtree below the selected node is removed to ensure semantic correctness in all situations.

\end{enumerate}

\begin{figure*}[h]
\begin{center}
\includegraphics[width=16cm]{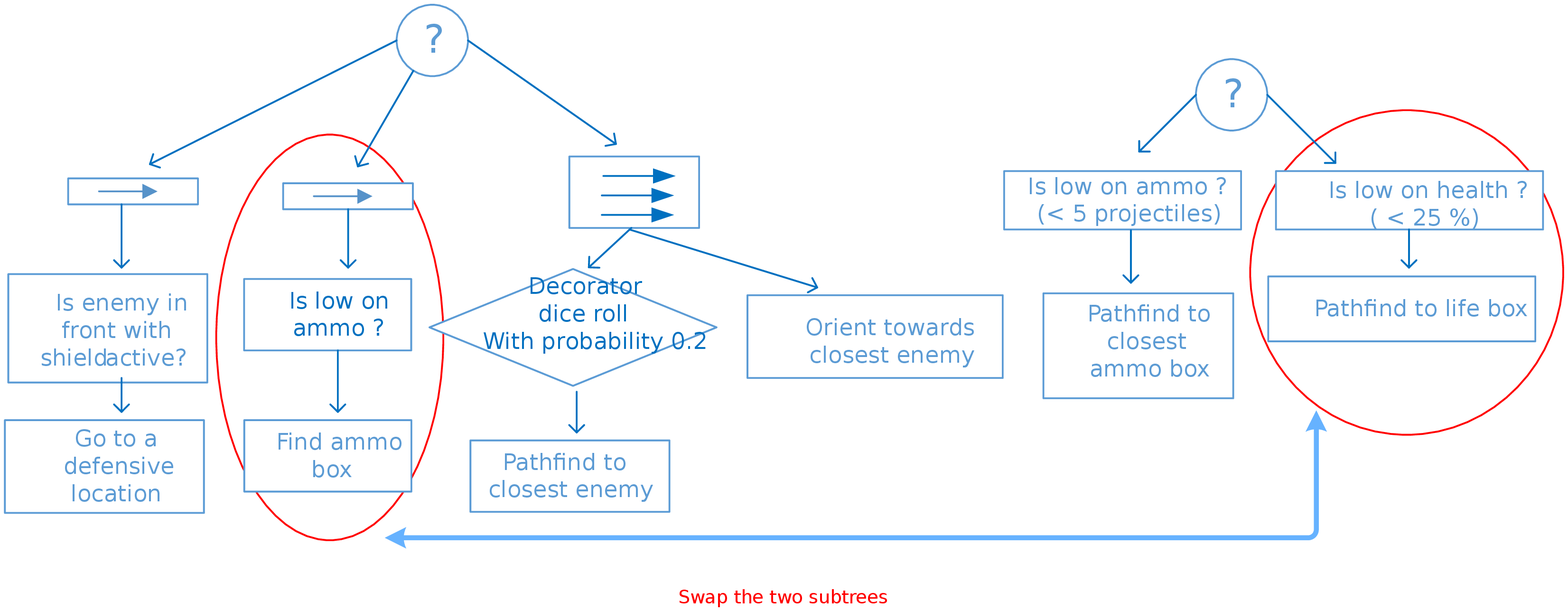}
\end{center}
\caption{Example of behavior tree exchanging subtrees to create new behaviors.}\label{fig:crossover}
\end{figure*}

\textbf{Cross-over}
The cross-over operator is exchanging subtrees between two different behavior trees (Figure \ref{fig:crossover}). The only observation about this operator is that the implementation takes care of invalid cases such as swapping subtrees and resulting in two decorators connected by an edge in one of the resulted behavior trees or performing garbage collection to remove recursively nodes with no children left.

\textbf{Selection}

The selection method in our implementation is a mix between Elitism \cite{Elitism}, Rank-based Roulette Wheel Selection \cite{srinivas1994genetic} and random selection, each one with a configurable percentage. 
The idea behind elitism is to keep between consecutive generations a given percentage of individuals that have the highest fitness values (candidates known as {\it elite}). This strategy is used since small modifications to a behavior tree can lead to very different fitness output, so preserving a part of the good examples might be a good solution. By $K$, we denote the number of individuals that must be preserved between consecutive generations. 

 The Rank-based Roulette Wheel Selection can add some variety by selecting probabilistically the individuals according to their rank in fitness value, while the random selection just adds some more variety of individuals by performing selection ignoring their fitness. Variable $EL$ represents the percentage of individuals selected with Elitism, $RW$ using the Rank-based Roulette Wheel Selection and $RR$ using random selection (i.e the constraint is to select these three variables such that $EL+RW+RR = 1$).

After the selection phase, mutation and crossover operations are applied to the $NL = N-K$ individuals left out of selection. Because at each new generation we keep the same number of individuals $N$, we create $NL$ new individuals, which are created using the ones unselected. 
The number of generations has an upper bound given as a parameter {\tt maxNumberOfGenerations}, but, before this limit is reached, a {\em plateau effect} is likely to be observed, i.e., when the individuals over a couple of generations do not improve significantly their fitness functions. More precisely, for two consecutive generations the change is checked as follows:
$$\sum_{x\in N} \big[\,\mbox{Fitness}\big(G_i[x]\big) - \mbox{Fitness}\big(G_{i-1}[x]\big)\,\big]^2 < \varepsilon.$$ 
One way to overcome this effect is to increase the parameters for mutation and cross-over, $P_{m}$ and $P_{c}$, temporarily until more diversity is added to the population of individuals \cite{burke1998putting}. In our implementation, if the average fitness is not improved in the last $NG$ generations, then $P_{m}$ and $P_{c}$ are gradually increased until they get to $P_{mMax}$ and $P_{cMax}$ after a specified number of generations. If the plateau still occurs, the algorithm is restarted because there is a low probability that it can find any better tests from this point. Figure \ref{alg:gen} shows the pseudocode of the framework's genetic algorithm.
\begin{figure}
\begin{tabular}{l} 
$G_0$ := population of $N$ individuals written \\
\ \ by a human user or randomly generated \\
\textbf{for} $i$ from $0$ to ${\tt maxNumberOfGenerations}$ \textbf{do}\\
\ \ \  $S_i$ := select $K$ individuals from $G_{i-1}$ \\
\ \ \  \ \ 	\ \ \ \ \	with probabilities.\ $EL$, $RW$, $RR$\\
\ \ \  $Others_i$  := generate $N-K$ individuals from $G_{i-1} \backslash S_i$\\
\ \ \  \ \ \  \ \ \  \ \ \  \ \ \  \ \
					using mutations and crossover \\
\ \ \  $G_i := S_i \cup Others_i$\\
\ \ \  Perform anti-bloating test \\
\ \ \  Check and add behavior trees to difficulty classes \\
\ \ \  Check for plateau and increase $P_{m}$ and $P_{c}$ if needed\\
\ \ \ \textbf{if} plateau still occurs after a number of generations \textbf{then} \\
\ \ \ \ \ \ STOP
\end{tabular}
\caption{Pseudocode of the genetic algorithm used inside framework implementation.}\label{alg:gen}
\end{figure}

In our experiments, we used $P_{m}=0.2$, $P_{c}=0.1$, $P_{mMax}=P_{m} + 0.3$, $P_{cMax}=P_{c} + 0.3$, $\varepsilon = 0.0001$, $K = N \times 0.25$, $EL = 0.75$, $RW = 0.15$, $RR = 0.1$. The maximum number of generations produced by the genetic algorithm {\tt maxNumberOfGenerations} was set to $50$. Similar values were reported as getting good results in the genetic algorithm's literature, but other methods such as particle swarm or machine learning can be used to find better values depending on the case.

The fitness functions are used as a dependency inversion paradigm by the framework, with users being able to inject their own evaluations of behavior trees. The framework is using out-of-the-box implementations that are presented in Section \ref{sec:difficulty}.

The anti-bloating technique \cite{article1} is also used in our approach to keep the dimensions of the generated behavior trees in reasonable dimensions (both in depth and width). At the end of each generation, for each tree one of the nodes located at the first or second level under root node is selected (probabilistically, with probability directly proportional to the depth of the subtree) and its entire subtree is eliminated. If the tree performance with respect to the used fitness function is nearly the same as the original,  the subtree elimination persists in the same generation, otherwise, the subtree is added back.

\subsection{Automatic difficulty management} \label{sec:difficulty}

Our Framework is able to create behavior trees automatically for different agent difficulty classes. The component \textit{Difficulty settings specifications} shown in Figure \ref{fig:communication}, defines the classes of difficulties that the user expects to be automatically created by the framework.
Formally, for each difficulty class, the user defines a set of metrics and their corresponding ranges (a $[Min-Max]$ interval of acceptable values) that are expected for that difficulty class and metric. (Eq. \ref{eq:diffmetrics}). Some of the metrics could overlap between different difficulty classes.

\begin{equation} \label{eq:diffmetrics}
DifficultyClass_{i}=\{Metric_{0}| Metric_{1}  | ....... | Metric_{N}\}.
\end{equation}

In our experiments we used metrics such as: the number of enemies destroyed, how many times an agent dies, distance traveled on map, how many times an agent escaped while its life percent was at some point smaller than $25\%$, how many times an agent collected a shield or weapon upgrade box, etc. 

The framework is trying to create a set of behavior trees for each of the difficulty classes in order. After each game round, feedback is received from the game side about these statistics for each of the agents. If the metrics are within the satisfiable limit of the class it is looking to generate behaviors, then it is added to that class difficulty set and the game simulation continues with the next generation until a fixed time stop.

At this level, the adaptive difficulty works as creating a custom class per human user, observing where the user is doing too bad or too worse (i.e. according to the metrics defined by the game), modify the corresponding metrics such that it offers a good experience for users automatically (i.e. growing or decreasing the $[Min-Max]$ range), then find a series of behavior trees that offer the right experience.

To create even more diversity, at the initialization of the population of behavior trees, each agent sets a fixed value for each of the metrics expected, instead of a range, i.e. a random value between $[Min-Max]$ for each $Metric_{i}$ is chosen and denoted further by $Target_{i}$. Then, our fitness function (Eq. \ref{eq:fitness}) tries to penalize agents that are not satisfying those target values set. This has the potential to bring even more diversity since, within each difficulty class, agents should still have various behaviors. Note that each metric target has its own weight ($w_{i}$ and all target and feedback values were normalized in $[0-1]$. A small $\varepsilon$ value to avoid division by 0.

\begin{equation} \label{eq:fitness}
Fitness = \frac{1}{\displaystyle\sum_{i=1}^{} w_{i}\times |\overline{Feedback_{i}}-\overline{Target_{i}}| + \varepsilon}
\end{equation}

\subsection{Behavior trees pruning} \label{sec:elimination}

While the genetic algorithm described above succeeded to produce enough behaviors in our evaluation, we noticed that there were a lot of similarities between them, and sometimes some of them failed to give the correct feedback when playing against a human user. To solve these issues we adopted a pipeline composed of three stages:
\begin{enumerate}

\item{A tournament selection method.} 

The main purpose of this stage is to eliminate behaviors that met the specified requirements just because their opponents were too weak (i.e. selection bias) in the generation phase. For a predefined number of rounds ($RT$) a random batch of $N$ behavior trees from each difficulty class is selected and $N$ agents are let to play against each other, each one using a unique tree from the selected set. At the end of each round, each tree received a score ($Score(tree, round)$) depending on their performance as defined by the same fitness function defined in Eq. \ref{eq:fitness}. At the end of all $RT$ rounds, the tournament score of each tree is defined as: 

\begin{equation} \label{eq:fitness}
TScore(tree) = \frac{\sum_{round}{} Score(tree, round)}{NumRoundsPlayed(tree)}
\end{equation}

If any of the behavior trees were not selected in any of the random batches, then it is removed to avoid zero division, but this did not happen during our tests by using a uniform distribution and considering a sufficient value for $RT$. Then, the average score per difficulty class is computed $AvgTScore(C)$ by simply averaging scores of behavior trees from each class. Finally, every behavior tree $T$ in a difficulty class rank $R$ which has a $TScore(T) < AvgTScore(R+1)$ is eliminated (i.e. if the behavior tree has an average score smaller than the average score of all trees in a lower rank difficulty class; the highest difficulty class is considered with rank 0). \\

\item{Similarity detection using K-means.}

With the output set of behavior trees surviving the tournament selection process, our method switches the focus on eliminating similar behaviors by using a K-means algorithm \cite{DBLP:journals/mst/BhattacharyaJK18}, and a custom similarity function (i.e. the distance between two generated trees) as defined in Equation \ref{eq:similarity}. The formula is a priority list of rules and scores, the first rule that applies is used.

\begin{equation} \label{eq:similarity}
\begin{split} 
Sim(N_{1},N_{2}) = 
     \begin{cases}
       \text{$N_{1}$ is sequence/selector and}  \\
            \text{\qquad \text{$N_{2}$ is not: 1.0}} \\
            \\
	   \text{$N_{1}$ is condition, $N_{2}$ is not:} \\
		    \text{\qquad\qquad \text{$0.2+Sim(N_{1}.child,N_{2})$}} \\
		    \\
	   \text{$N_{1}$ and $N_{2}$ are conditions:}  \\
	        \text{\qquad$ParamsDiff(N_{1},N_{2})$} \\
		    \text{\qquad\qquad\text{$+Sim(N_{1}.child,N_{2}.child)$}} \\
		    \\
	   \text{if $N_{1}$ and $N_{2}$ are both action nodes:} \\
	   		\text{\qquad \text{if they are different actions: 0.2}} \\
	   		\text{\qquad\text{else: 0.0}} \\
	   		\\
	   \text{If both $N_{1}$ and $N_{2}$ have children:} \\
			\text{\qquad\text{$0.0+Sim(N_{1}.child,N_{2}.child)$}} \\
			\\
	   \text{else: 0.0} \\
     \end{cases}
\end{split} 
\end{equation}

The K-means algorithm can be executed to group the behaviors inside each difficulty class, within K clusters that are different in relation to the custom similarity metric defined. Finally, a member of each of these clusters is selected. Thus, the K value represents how many behaviors are actually needed by the game at runtime for each difficulty class. \\

\item{Human feedback}

Concretely, human players from an internal \text{Quality Assurance} department are let to play against AI agents using random behavior trees for a limited time. Their role is to vote the agents with unnatural behaviors while playing the game. This stage is optional, but it can be efficient in identifying some of the behavior trees that don't look natural, since evaluating the human-likeness of agents is a very difficult task to formalize and automatize. 

\end{enumerate}

\subsection{Automated functional testing}

Code coverage is one of the most used metrics in evaluating how much source code was tested against defects \cite{Ammann:2008:IST:1355340}. Typical automated functional tests are written in games to test the source code. Behavior trees can be seen as such functional tests indeed since they connect multiple systems in the same test execution in a way also known in the literature as model based testing.  

However, the code coverage metric shows only the number of lines of code touched during the test process. While this can be a valuable thing in general, it would be ideal to test the same (or more) number of lines of source code in various orders, or against different input values. Behavior trees can be useful for this purpose since they can connect multiple systems in different orders and with different parameters. More details are presented in the evaluation section.

\section{Evaluation} \label{sec:evaluation}
The environment setup was already described in Section \ref{sec:optimizations}. 
A single round took in average 5 minutes to finish, until a single agent out of an initial number of 8 agents remained the map, then restart. Between rounds, the initial positions of the agents were randomized to ensure that there is no advantage in one or another. At a first look, faster gameplay simulation can be done by fictively increasing the game speed (in our case the speed of agents moving around and projectiles), thus increasing the chance of getting more and more behaviors with fewer resources consumed. However, from our experiments, increasing these speeds too much over the normal game speed can make behaviors exploit the timing issues and generalize for that faster gameplay only.

\subsection{Genetic algorithm performance}
The generation algorithm and parameter values were detailed in Section \ref{sec:rules}. Below we explain the performance of the results along with the techniques that helped us cut some of the behaviors that looked as being the right category after generation, but weren't. The results were observed after leaving the simulation running for $24h$ and averaging the results in $5$ runs. A single computer was used for the simulation with the following specifications: Intel Core i7 8600K CPU, 8 GB RAM and, an Nvidia 1060 GTX video card. The framerate of the simulation was capped at $60$ ticks per second.

\textit{Generation of trees per class}
Table 1 shows the number of behavior trees eligible for each difficulty class, according to the metrics specified before, by using a single process executing the game instance. 

\begin{table}[htbp]
\label{t:results1}
\caption{Table showing the total number of behaviors generated by averaging 5 runs each lasting a fixed time of $24h$, and the number of different behaviors found for each difficulty class in our simulation according to the defined metrics.}
\begin{center}
\begin{tabular}{|c|c|}
\hline
 Number of BTs created & 5421 \\ 
\hline
 Easy & 129 \\ 
\hline
 Medium & 46 \\ 
\hline
 Hard & 17 \\ 
\hline
\end{tabular}
\label{tab1}
\end{center}
\end{table}

The number of behavior trees found per difficulty class is affected by the user defined metrics, and also by the availability of the actions and conditions offered by the game implementation. The way we did tuning in the experiments was to understand first what should a smarter or weaker agent do, and create more actions and conditions to cover those behaviors. Using the anti-bloating technique described in \ref{sec:rules}, we managed to keep the generated trees in an understandable format for a human viewer: depths were between 4-9, while width in the range 3 to 8. Tables \ref{tab2} and \ref{tab3} show what happened to the total of 5421 trees evaluated during our experiment. The remaining of $4.4\%$ from the initial proposed set of BTs were classified as one of the three difficulty classes according to the user's specification (Table \ref{tab1}). The tournament evaluation was allowed to execute for another $48h$ in continuation of the generation phase. The $K$ parameter for similarity grouping inside each difficulty class can be set more aggressively in environments were behaviors variety and their parameters range are limited. 

\begin{table}[htbp]
\caption{Showing how code issues, metrics imposed and tournament evaluation affected the initial set of proposed behavior trees.}
\begin{center}
\begin{tabular}{|p{6cm}|c|}
\hline
 Total number of BTs created & 5421 \\ 
\hline
 Total number of unclassified BTs & 5229 ($95.6\%$) \\ 
\hline
 Number of BTs that crashed or encountered other issues that prevented a evaluation & 31 ($0.6\%$) \\ 
\hline
 Number of BTs that didn't pass the quality metrics for any of the three pre-defined classes & 2474 ($45.4\%$) \\
\hline
 Number of BTs rejected by the tournament evaluation & 1237 ($22.6\%$) \\
\hline
\end{tabular}
\label{tab2}
\end{center}
\end{table}

\begin{table}[htbp]
\caption{Showing similarity and human reporting results.}
\begin{center}
\begin{tabular}{|p{6cm}|c|}
\hline
 Remaining number of BTs after prunning defined in Table \ref{tab1} & 1679 \\
\hline
 Parameters $K$ for K-means algorithm used in similarity detection Easy, Medium and Hard difficulty classes & 150 | 50 | 30 \\ 
\hline
 Number of behavior trees voted as not natural by human users per class & 21 | 4 | 13 \\
\hline
\end{tabular}
\label{tab3}
\end{center}
\end{table}
 
\textit{Transfer learning}
In our experiments, letting the framework generate the initial set of behavior trees almost doubled the time needed to obtain the same results. An important benefit of the approach is that even after adding new actions and conditions to the set, the behavior trees can evolve quickly from the previous set of learned behaviors which used the old sets (i.e. similar to the concept of transfer learning \cite{6729500}). If actions and conditions are removed from the sets, their subtrees are completely deleted from the existing sets, using the same rules defined in Section \ref{sec:rules}.

\begin{figure*}[t]
\begin{center}
\includegraphics[width=12cm]{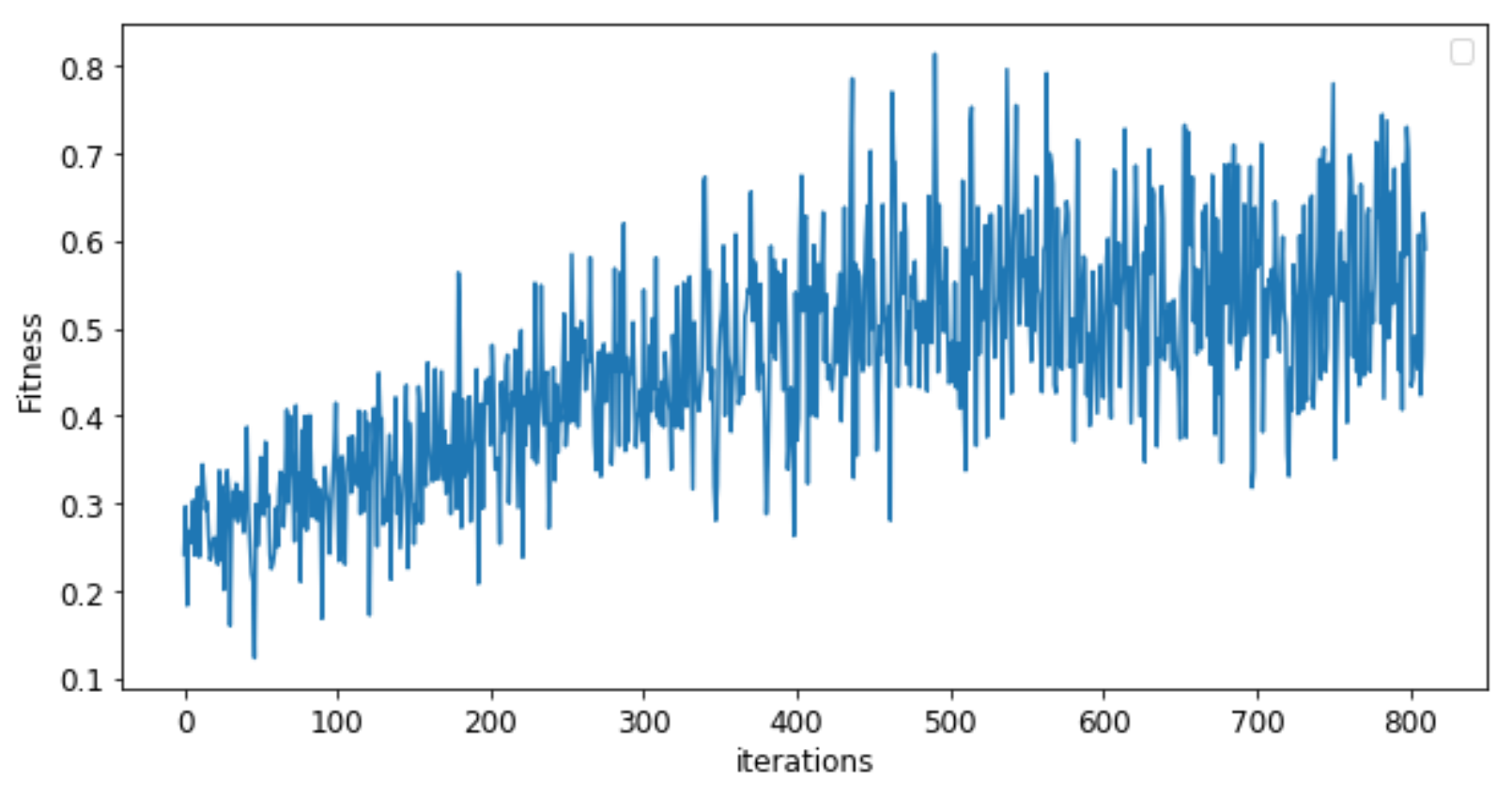}
\end{center}
\caption{A graph showing the min-max range of the fitness values in our experiment for all iterations.}\label{fig:fitness}
\end{figure*}

\textit{Fitness per time}
Figure \ref{fig:fitness} shows the evolution of the fitness values in our experiment during all 811 simulated rounds in the $24h$ interval. It is important to notice that the average fitness improves over time (especially in the first half of the rounds), while the variance shows that it is possible to get \textit{easy} class difficulty behaviors at the end of the simulation when the average fitness is high, or \textit{hard} class difficulty behaviors in the middle of the simulation.

\subsection{Testability and possible exploits reporting}
In our experiments, we built a set of $20$ behavior trees by hand covering all sets of actions, conditions, and decorators, such that the code coverage metric for the gameplay components implementation was nearly $100\%$, and no remaining defects were shown in $5$ rounds of the game. Then, we considered the same set as the initial population for the genetic algorithm (with the fitness function used for the hard class difficulty, but this aspect shouldn't matter in general).  After $12h$ of running time of automated tests running on the same machine, we found $19$ more defects in the code caused by memory corruptions (running over the same lines of code but with different inputs was really helpful), multithreading or determinism breaking issues (i.e. running multiple times with the same initial state and behaviors attached to agents resulted in a different game state after some time) - these were discovered especially because of the different orders in which certain components were accessed, which initially looked quite unnatural. This is an important aspect since many times when building behaviors hierarchically, ideally they should be reused without concerns of what is inside the building blocks a user is connecting.

A formal reporting of behaviors that look more like exploits, which takes less human effort to recognize, is on the future work list. One example of such exploit found was that of a behavior created by the framework where the agent was just looking for shield boxes, stayed in cover and firing against the closest enemy. This behavior was the most efficient one at an early point, being included in the Hard difficulty class. To fix this, we decreased the time that shield is active and made less shield boxes available on the field. This forced the agents to try other various strategies. The only automated reporting available at this point is a basic formula involving the width and height of the behavior tree. However, this can be easily fooled by the framework, which can create some fictive conditions that make the width or height bigger, but still having a basic exploiting behavior.

\section{Conclusion and future work} \label{sec:conclusion}
The paper presented the ideas and implementation details behind a framework which can be used for games and simulation environments for creating diverse behaviors, automated and adaptive difficulties, and automated functional test of defects and exploits. The framework uses at the moment genetic algorithms and obtained with limited computational resources enough variety of behaviors in a short time. 

While the motivating problems and solutions in the paper are presented from a video game perspective, we think that these can be adapted to many other types of applications that involve simulations of virtual environments with multi-agent systems.

Our future plans will focus on integrating behavior trees with machine learning methods (especially reinforcement learning) from two different perspectives: finding policies for each difficulty class, and also study the mechanisms for creating behavior tree nodes that are applying those policies together with flow management mechanisms.

\section{\uppercase{ACKNOWLEDGMENTS}}

\noindent 
This work was supported by a grant of Romanian Ministry of Research and Innovation CCCDI-UEFISCDI. project no. 17PCCDI/2018.

\bibliographystyle{splncs03}
\nocite{*}
\bibliography{conference_041818}

\end{document}